\documentclass[letterpaper]{article} 
\usepackage{aaai2026}  
\usepackage{times}  
\usepackage{helvet}  
\usepackage{courier}  
\usepackage[hyphens]{url}  
\usepackage{graphicx} 
\usepackage{dsfont}
\usepackage{natbib}  
\usepackage{caption} 
\usepackage{algorithm}
\usepackage{algorithmic}

\usepackage{newfloat}
\usepackage{listings}
\DeclareCaptionStyle{ruled}{labelfont=normalfont,labelsep=colon,strut=off} 
\floatstyle{ruled}
\newfloat{listing}{tb}{lst}{}
\floatname{listing}{Listing}
\usepackage{xspace}
\usepackage{booktabs}       
\usepackage{amsfonts}       
\usepackage{nicefrac}       
\usepackage{microtype}      
\usepackage{xcolor}         
\usepackage{amsmath}        
\newcommand{\model}{RTMol\xspace}
\newcommand{\arrow}[1]{\,\scalebox{1.0}{#1}}
\newcommand{\up}{\arrow{$\uparrow$}}

\title{\model: Rethinking Molecule-text Alignment in a Round-trip View}

\author{
  Letian Chen\equalcontrib\textsuperscript{\rm 1, \rm 2}, 
  Runhan Shi\equalcontrib\textsuperscript{\rm 2}, 
  Gufeng Yu\textsuperscript{\rm 2}, 
  Yang Yang\textsuperscript{\rm 2}\thanks{Corresponding author.}
}
\affiliations {
    \textsuperscript{\rm 1}Shanghai Innovation Institute\\
    \textsuperscript{\rm 2}School of Computer Science, Shanghai Jiao Tong University\\
    \texttt{\{clt2001, han.run.jiangming, jm5820zz\}@sjtu.edu.cn}, \texttt{yangyang@cs.sjtu.edu.cn}
}

\begin{document}

\maketitle

\begin{abstract}

Aligning molecular sequence representations (e.g., SMILES notations) with textual descriptions is critical for applications spanning drug discovery, materials design, and automated chemical literature analysis. Existing methodologies typically treat molecular captioning (molecule-to-text) and text-based molecular design (text-to-molecule) as separate tasks, relying on supervised fine-tuning or contrastive learning pipelines. These approaches face three key limitations: (i) conventional metrics like BLEU prioritize linguistic fluency over chemical accuracy, (ii) training datasets frequently contain chemically ambiguous narratives with incomplete specifications, and (iii) independent optimization of generation directions leads to bidirectional inconsistency. To address these issues, we propose RTMol, a bidirectional alignment framework that unifies molecular captioning and text-to-SMILES generation through self-supervised round-trip learning. The framework introduces novel round-trip evaluation metrics and enables unsupervised training for molecular captioning without requiring paired molecule-text corpora. Experiments demonstrate that RTMol enhances bidirectional alignment performance by up to 47\% across various LLMs, establishing an effective paradigm for joint molecule-text understanding and generation.
\end{abstract}

\begin{links}
    \link{Code}{https://github.com/clt20011110/RTMol}
\end{links}

\section{Introduction}
Understanding molecular sequence representations and enabling \textit{de novo} molecular design constitute fundamental challenges in cheminformatics and computational chemistry~\cite{Mouchlis2021AdvancesID}. 
With molecules expressible as linear strings through the Simplified Molecular Input Line Entry System (SMILES)~\cite{smiles}, recent studies have employed large language models (LLMs) for molecular understanding and generation~\cite{li2024molreflect,zhang2024unimot,Sadeghi2024CanLL}. These models leverage vast biochemical knowledge embedded in scientific literature and hold promise for unifying representation learning across textual descriptions and molecular structural domains. (See Appendix~A for more related work.)

\begin{figure}
    \centering
    \includegraphics[width=0.95\linewidth]{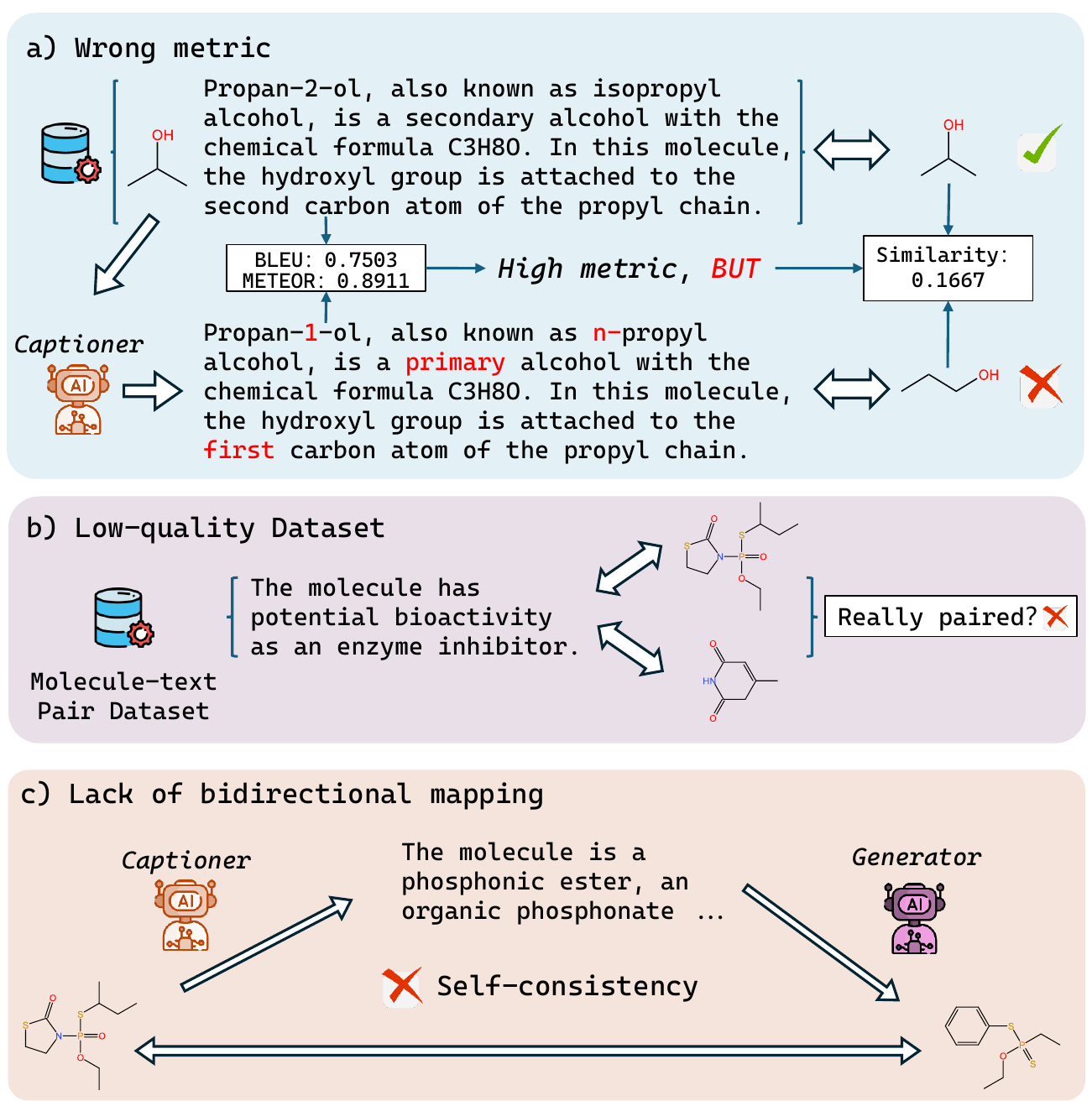}
    \caption{Limitations of current molecule-text alignment: (a) textual metrics ignore chemical fidelity; (b) captions are noisy or incomplete; and (c) separate modeling fails to enforce bidirectional consistency.}
    \label{fig:drawback}
\end{figure}

The critical challenge resides in establishing bidirectional alignment between molecular sequences and textual descriptions, a prerequisite for robust cross-modal reasoning. Current methodologies approach this through two decoupled tasks: i) Molecular captioning (molecule-to-text), and ii) Text-based molecular design (text-to-molecule). Although conceptually inverse, existing frameworks optimize these objectives independently through separate training regimes. Moreover, the prevailing reliance of current methodologies on supervised learning creates a dependency on high-quality paired datasets and robust evaluation metrics. As illustrated in Fig.~\ref{fig:drawback}, this paradigm introduces three fundamental limitations.

First, evaluation standards for molecular captioning are often misleading. Metrics such as BLEU~\cite{bleu} and METEOR~\cite{meteor}, which are commonly used to assess generated texts, primarily measure textual similarity through $n$-gram overlap. While these metrics reward fluent and keyword-rich captions, they often ignore chemical factuality. As a result, captions that achieve high BLEU/METEOR scores may still misrepresent molecular structures, containing incorrect or misleading chemical details. Recent analyses, such as those in molecular captioning benchmarks, demonstrate that high-scoring captions may violate chemical facts, indicating a weak alignment between textual quality and chemical correctness~\cite{guo2023can}.

Second, the quality of existing datasets is problematic. Many publicly available molecule-text datasets suffer from ambiguous or incomplete descriptions~\cite{molt5,drawbackDataset}. These descriptions are often generic, failing to uniquely identify the associated molecule or to describe key structural features. Using such datasets for supervised learning in molecular captioning tasks impairs the model's ability to generalize reliably during inference. 

Third, the lack of bidirectional alignment causes inconsistent understanding. Current training treats molecular captioning and text-based molecular design as separate tasks. Consequently, a model proficient in one often fails when its output is used as input for the other, leading to fragmented knowledge of molecule-text relationships~\cite{guo2023can}. This inconsistency is a key barrier to unified modeling. Furthermore, achieving this bidirectional alignment is significantly hindered by the two previously mentioned limitations: the absence of robust evaluation metrics and large-scale, high-quality datasets.

To address these challenges, we propose a round-trip learning framework that unifies molecular captioning and text-based molecular design into a single training paradigm. The framework encourages round-trip consistency: the model first generates a caption from a molecule and then reconstructs the molecule solely from that caption. The similarity between the original and reconstructed molecules serves as a reward signal, directly optimizing for chemically faithful descriptions and bypassing reliance on noisy labels or purely textual metrics. This integrated approach promotes robust bidirectional alignment between molecular and textual representations. Our main contributions are:
\begin{itemize}
\item We propose a round-trip metric that evaluates captions by chemical fidelity rather than textual overlap, overcoming the limitations of conventional metrics.

\item We present \model, a round-trip reinforcement learning framework that jointly aligns molecular captioning and text-based molecule generation.

\item Our framework enables self-supervised training for molecular captioning, reducing reliance on noisy or incomplete annotations.

\item Experiments demonstrate consistent improvements in both round-trip captioning and text-to-molecule tasks across various LLM backbones.

\end{itemize}

\begin{figure*}
    \centering
    \includegraphics[width=0.95\linewidth]{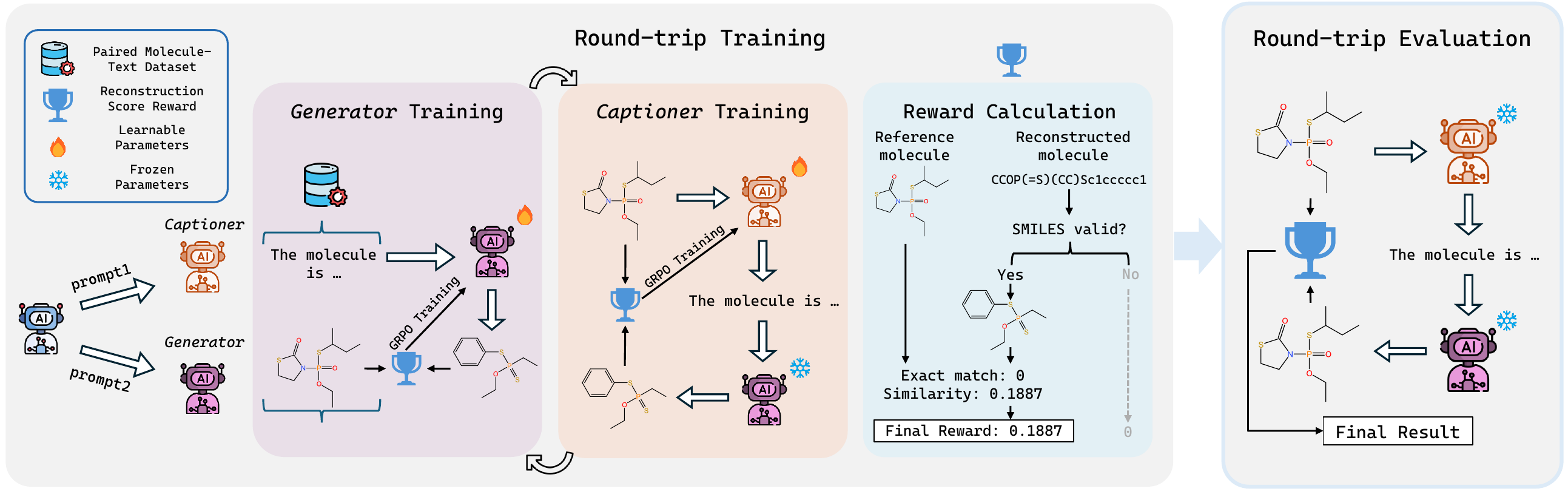}
    \caption{Overview of \model. A single LLM serves as both the Captioner and Generator for molecule-text alignment, with their training alternating cyclically in a complementary manner to reinforce each other's performance.}
    \label{fig:framework}
\end{figure*}

\section{Method}

\subsection{Problem Formulation\label{sec:pf}}

Let $\mathcal{M}$ denote the space of molecules and $\mathcal{T}$ the space of text descriptions. We consider two conditional probability distributions:
\begin{itemize}
    \item $p_{\theta}(y|x)$, which models the probability of generating a textual description $y \in \mathcal{T}$ given a molecule $x \in \mathcal{M}$.
    \item $q_{\phi}(x'|y)$, which models the probability of generating a molecule $x' \in \mathcal{M}$ given a text description $y \in \mathcal{T}$.
\end{itemize}

The round-trip process involves first sampling text $y \sim p_{\theta}(y|x)$ given the original molecule $x$, and then sampling a reconstructed molecule $x' \sim q_{\phi}(x'|y)$.

To encourage faithful alignment between the molecular and textual modalities, we aim to minimize the expected discrepancy between the original molecule $x$ and its reconstruction $x'$. This discrepancy is quantified by a distance function $d: \mathcal{M} \times \mathcal{M} \rightarrow \mathbb{R}_{\ge 0}$, which measures the structural difference between two molecules.

Formally, the goal is to learn optimal parameters $\theta^*$ and $\phi^*$ such that the expected reconstruction loss over the joint distribution of $(x, y, x')$ is minimized. The sampling procedure follows $x \sim p(x)$, $y \sim p_{\theta}(y|x)$, $x' \sim q_{\phi}(x'|y)$, where $p(x)$ denotes the empirical distribution of molecules in the dataset. The optimization objective is defined as:
\begin{equation}\label{eq:obj}
    (\theta^*, \phi^*) = \arg\min_{\theta, \phi} \mathbb{E}_{x \sim p(x),y \sim p_{\theta}(y|x),x' \sim q_{\phi}(x'|y)} \left[ d(x, x') \right].
\end{equation}
This objective aims to learn a pair of conditional distributions that effectively align the molecular space $\mathcal{M}$ and the textual space $\mathcal{T}$, by minimizing the loss incurred during the round-trip process from a molecule to text and back to a reconstructed molecule.

\subsection{Round-trip Metric}
To evaluate the quality of molecule-text alignment, we introduce the round-trip metric $R$. The metric measures how well the model preserves information after translating the molecule into text and reconstructing it back into molecular form. The metric is defined as, 

\begin{equation}\label{eq:rt_0}
    R(\theta, \phi) = \mathbb{E}_{x\sim p(x),y \sim p_{\theta}(y|x),x' \sim q_{\phi}(x'|y)}[\mathds{1}\{d(x,x')=0\}],
\end{equation}
where 
$\mathds{1}$ is a boolean indicator function. A higher $R$ means a higher expectation that the reconstructed molecule $x'$ is the same as the original molecule $x$, indicating better round-trip fidelity and stronger alignment between the two modalities.

\subsection{Round-trip Optimization}
We now show that 
minimizing Eq.\ref{eq:obj}'s objective is equivalent to maximizing a variational lower bound on the molecular-textual mutual information.

Let $X\in\mathcal{M}$ denote a random molecule drawn from $p(x)$,
and let $Y\in\mathcal{T}$ be the text description corresponding to $X$,
i.e.\ $Y\mid X\sim p_{\theta}(y\mid X)$.
Viewing $X$ and $Y$ as two random variables, the mutual information between $X$ and $Y$ is
\begin{equation}\label{eq:mi}
   I(X,Y;\theta)=\mathbb{E}_{p(x)\,p_\theta(y|x)}
          \!\bigl[ \log p_\theta(y|x)-\log p(y) \bigr].
\end{equation}
Using the Barber-Agakov variational decomposition~\cite{barber}, Eq.~\ref{eq:mi} becomes
\begin{align}
   I(X,Y;\theta)&=H(X)+\mathbb{E}_{p(x)\,p_\theta(y|x)}\!
         \bigl[\log q_\phi(x|y)\bigr]       \notag\\
      &\hspace{2em}
        +\mathrm{KL}(p(x|y)\,\Vert\,q_\phi(x|y)) \notag\\
        &\;\;\ge\;H(X)+\mathbb{E}_{p(x)\,p_\theta(y|x)}
         \!\bigl[\log q_\phi(x|y)\bigr],
   \label{eq:BA}
\end{align}
where $H(X)$ is the entropy of $X$ and KL($\cdot\Vert\cdot$) denotes the Kullback-Leibler divergence. 
If $q_\phi$ is $L$-Lipschitz with respect to the distance function $d$ defined above, then for some $\alpha>0$, we have
\begin{equation}
   \log q_\phi(x|y)\;\;\ge\; \mathbb{E}_{\,q_\phi(x'|y)}\bigl[-\alpha\,d(x,x')+C\bigr],
   \label{eq:lipschitz}
\end{equation}
where $C$ is a constant (proof in Appendix~B).  
Substituting Eq.~\ref{eq:lipschitz} into Eq.~\ref{eq:BA} yields
\begin{equation}
   I(X,Y;\theta)\;\;\ge\;H(X)-\alpha\,
      \mathbb{E}_{p(x)\,p_\theta(y|x)\,q_\phi(x'|y)}\bigl[d(x,x')\bigr]+C,
   \label{eq:mi_lower}
\end{equation}
which concludes the proof. Higher round-trip metrics correspond to smaller distances between original and reconstructed molecules, creating a theoretically grounded metric through its direct relationship with the distance function. 

\subsection{Model Architecture}

Our framework, as shown in Fig.~\ref{fig:framework}, uses a single large language model (LLM) that operates in two complementary roles: 
\begin{itemize}
    \item \textbf{Generator} ($q_\phi(x|y)$): converts textual descriptions into molecules.
    \item \textbf{Captioner} ($p_{\theta}(y|x)$): generates textual descriptions from molecules.
\end{itemize}
Unlike traditional methods that train these tasks independently, we couple them in a unified training process to establish consistent bidirectional alignment between molecules and text.

\paragraph{Generator: Grounding in Chemical Knowledge.}

The Generator translates natural language descriptions into valid molecular SMILES. It is trained using supervised fine-tuning on paired molecule-text data and optimized to maximize the reconstruction score $S$. The score evaluates three aspects:
\begin{itemize}
    \item Similarity $S_{\text{sim}}(\cdot, \cdot)$: the similarity of molecular fingerprints between the original and reconstructed molecules.
    \item Exact match $S_{\text{exact}}(\cdot, \cdot)$: whether the reconstructed SMILES exactly matches the original.
    \item Validity $S_{\text{valid}}(\cdot)$: whether the generated SMILES string represents a chemically valid molecule.
\end{itemize}
Formally, the reconstruction score is defined as, 
\begin{equation}\label{eq:rt}
\begin{aligned}
    S(x, x') &= 
    \begin{cases}
    0,& \text{if } S_{\text{valid}}(x')=0, \\
    S_{\text{sim}}(x, x') + S_{\text{exact}}(x, x'),& \text{otherwise}, \\
    \end{cases}
\end{aligned}
\end{equation}
where
\begin{equation}\label{eq:metric}
\begin{aligned}
    S_{\text{valid}}(x') &= \mathds{1}\{x' \text{ is valid}\},\\
    S_{\text{exact}}(x, x') &= \mathds{1}\{x=x'\},\\
    S_{\text{sim}}(x, x') &= T_\mathrm{MACCS}(x, x') + T_\mathrm{RDKit}(x, x') \\
    &\quad\ + T_\mathrm{Morgan}(x, x').
\end{aligned}
\end{equation}
Here each $T_f(x, x')$ denotes the Tanimoto coefficient~\cite{Bajusz2015b} computed using a specific chemical fingerprint representation: 
\begin{align}
T_f(x, x') = \frac{|f(x) \cap f(x')|}{|f(x) \cup f(x')|},
\end{align}

where $f \in \{\mathrm{MACCS}, \mathrm{RDKit}, \mathrm{Morgan}\}:\mathcal{M}\rightarrow 2^{\{0,1\ldots,n\}} $ is a fingerprint function that maps a molecule to a set of integer feature identifiers. These functions represent molecular structures through MACCS keys~\cite{maccs}, RDKit (RDK) path-based fingerprints~\cite{rdkit}, and Morgan circular fingerprints~\cite{morgan}, respectively. 
We combine three molecular fingerprints to capture complementary chemical features, including substructures, physicochemical properties, and topological scaffolds.
This multi-fingerprint similarity metric grounds the Generator's understanding in established chemical representations, enabling it to serve as a reliable evaluator.

\paragraph{Captioner: Learning via Round-trip Process.}
The Captioner learns to describe a molecule with natural language. Critically, its training is \textbf{unsupervised} concerning textual labels: instead of matching reference captions, the Captioner learns through a round-trip interaction with the Generator. For a given molecule $x$, the Captioner first samples a description $y\sim p_{\theta}(y|x)$, which is immediately passed to the Generator to reconstruct a molecule $x'\sim q_{\phi}(x'|y)$. The Captioner's objective is to maximize the reconstruction score defined in Eq.~\ref{eq:rt} between $x$ and $x'$, directly rewarding captions that preserve chemically relevant information. This self-consistency training bypasses the limitations of ambiguous or low-quality text labels, driving the model to produce descriptions that are both precise and informative for reconstruction. 

\subsection{Training Strategy}

\paragraph{Choice of Optimization Algorithm.}
We optimize our model using reinforcement learning (RL) due to the non-differentiable nature of our objective. The final reconstruction score $S$ defined in Eq.~\ref{eq:rt} involves components such as chemical validity checks and exact SMILES string matches, which cannot be optimized with standard backpropagation. We therefore frame the task as an RL problem, where the Captioner acts as a policy network. It generates textual descriptions (actions) to maximize an expected reward signal. For the optimization itself, we employ the Group Relative Policy Optimization (GRPO) (see Appendix~C), which is well-suited for this rule-based reward setup.

\paragraph{Choice of Reward Function.}
Directly using the final round-trip metric $R(\theta,\phi)$ defined in Eq.~\ref{eq:rt_0} as the sole reward is ineffective, particularly in the early stages of training. The initial, untrained Captioner backbone frequently generates syntactically invalid SMILES strings and produces captions that result in a reconstruction score of zero. Such a sparse and uninformative reward signal fails to provide a meaningful gradient for optimization. As shown in Eq.~\ref{eq:rt}, we include validity checking to enforce the Captioner to generate valid SMILES strings and a similarity reward to make the reward landscape smoother, providing a continuous optimization gradient even when an exact molecular match is not achieved.

\paragraph{Coupled Training.}

The Captioner and Generator are trained in a closed-loop process. Crucially, the Generator acts as an evaluator for the Captioner’s output, creating a dependency: the Captioner’s improvement is directly dependent on the Generator’s feedback. While the Generator’s training remains independent of the Captioner’s outputs, the two models are trained in parallel to prevent degradation of the Generator’s core text-to-molecule capability. This ensures the Captioner consistently learns from a stable, proficient evaluator, which is essential for refining molecular-textual alignment. The training pseudo-code is provided in Appendix~D.

\section{Experiments}

\subsection{Experimental Setup}

\paragraph{Dataset.}

Our primary experiments use the ChEBI-20 dataset~\cite{edwards2021text2mol} (33,010 molecule-text pairs), with a standard 8:1:1 random split for training, validation, and testing. 
For fair comparison, main evaluations follow prior work by using a 100-sample benchmark test set~\cite{zhao2024chemdfm}.
We additionally evaluate on two external datasets, L+M~\cite{LM} and Mol-Instruction~\cite{mol_instruct}. While Mol-Instruction shows high overlap with ChEBI-20 (88.9\%), L+M suffers from underspecified textual descriptions. To prevent data leakage, we remove overlapping samples, yielding filtered versions L+M-F (180,178 pairs) and Mol-Instruct-F (239,659 pairs).

\paragraph{Baselines.}

We benchmark our model against two categories of baselines: 
1) {General-purpose LLMs}: GPT-4o~\cite{openai2024gpt4}, Gemini-2.5-Flash~\cite{Comanici2025Gemini2P}, Qwen-3-8B~\cite{Yang2025Qwen3TR}, and DeepSeek-V3~\cite{DeepSeekAI2024DeepSeekV3TR}; and 
2) {Domain-specific models}: ChemT5-0.2B~\cite{chemt5}, ChemDFM-8B~\cite{zhao2024chemdfm}, and ether0~\cite{ether0}.  
{For the \model framework}, we implement three variants using Qwen-3-8B (Qwen3), ChemT5-0.2B (ChemT5), and ChemDFM-8B (ChemDFM) as base models.

\begin{table*}[h!]
\centering
\begin{tabular}{@{}lccccccc@{}}
\toprule
\textbf{Model} & \textbf{Exact(\%)}\up & \textbf{Validity(\%)}\up & \textbf{MACCS}\up & \textbf{RDK}\up & \textbf{Morgan}\up & \textbf{BLEU}\up & \textbf{METEOR}\up \\
\midrule
\multicolumn{1}{l}{\textit{Baselines}} & \multicolumn{5}{c}{\textit{Chemical metrics}} & \multicolumn{2}{c}{\textit{Textual metrics}} \\
\midrule
GPT-4o & 3.0  & 72.0  & 0.541  & 0.384  & 0.209  & 0.425  & 0.548  \\
DeepSeek-V3    & 3.0  & 85.0  & 0.704  & 0.521  & 0.343  & 0.575  & 0.694  \\
Gemini-2.5-Flash    & 17.0  & 40.0  & 0.374  & 0.343  & 0.293  & 0.597  & 0.565  \\
ether0    & 4.0  & 71.0  & 0.385  & 0.291  & 0.198  & 0.280  & 0.405 \\
\midrule
\multicolumn{8}{l}{\textit{Round-trip training}} \\
\midrule
Qwen3     & 7.0 & 60.0  & 0.409  & 0.326  & 0.234  & 0.392  & 0.569 \\
Qwen3\,+\,\model   & \underline{9.0} & \underline{92.0}  & \underline{0.580} & \underline{0.409} & \underline{0.274} & \underline{0.465} & \underline{0.581} \\
\midrule
ChemT5     & 12.0 & 86.0  & 0.691  & 0.605  & 0.482  & \underline{0.595}  & 0.695  \\
ChemT5\,+\,\model   & \underline{14.0} & \underline{87.0}  & \underline{0.699} & \underline{0.613} & \underline{0.486} & 0.590 & \underline{0.701} \\
\midrule
ChemDFM     & 19.0 & 90.0  & 0.669  & 0.579  & 0.457  & 0.603  & 0.734 \\
ChemDFM\,+\,\model   & \underline{28.0} & \underline{98.0}  & \underline{0.826} & \underline{0.729} & \underline{0.597} & \underline{0.722} & \underline{0.812} \\
\bottomrule
\end{tabular}
\caption{Benchmark results of different models in round-trip evaluation using generated descriptions. The best results are highlighted in underline for each model.}
\label{tab:round-trip}
\end{table*}

\begin{table*}[h!]
\centering
\begin{tabular}{@{}lccccccc@{}}
\toprule
\textbf{Model} & \textbf{Exact(\%)}\up & \textbf{Validity(\%)}\up & \textbf{MACCS}\up & \textbf{RDK}\up & \textbf{Morgan}\up & \textbf{BLEU}\up & \textbf{METEOR}\up \\
\midrule
\multicolumn{1}{l}{\textit{Baselines}} & \multicolumn{5}{c}{\textit{Chemical metrics}} & \multicolumn{2}{c}{\textit{Textual metrics}} \\
\midrule
GPT-4o         & 5.0  & 75.0 & 0.593 & 0.432 & 0.277 & 0.427 & 0.555 \\
DeepSeek-V3    & 18.0 & 86.0 & 0.764 & 0.618 & 0.468 & 0.530 & 0.674 \\
Gemini-2.5-Flash & 16.0 & 69.0 & 0.635 & 0.551 & 0.441 & 0.605 & 0.662 \\
ether0         & 5.0  & 28.0 & 0.165 & 0.126 & 0.096 & 0.121 & 0.234 \\
\midrule
\multicolumn{8}{l}{\textit{Round-trip training}} \\
\midrule
Qwen3       & 2.0  & 53.0 & 0.290 & 0.180 & 0.106 & 0.187 & 0.432 \\
Qwen3\,+\,\model  & \underline{3.0}  & \underline{88.0} & \underline{0.469} & \underline{0.281} & \underline{0.142} & \underline{0.283} & \underline{0.465} \\
\midrule
ChemT5     & 12.0 & 89.0 & 0.753 & 0.636 & 0.508 & \underline{0.519} & 0.691 \\
ChemT5\,+\,\model  & \underline{13.0} & \underline{90.0} & \underline{0.765} & \underline{0.648} & \underline{0.524} & 0.518 & \underline{0.694} \\
\midrule
ChemDFM      & 24.0 & 96.0 & 0.864 & 0.755 & 0.620 & 0.797 & 0.903 \\
ChemDFM\,+\,\model  & \underline{54.0} & \underline{99.0} & \underline{0.936} & \underline{0.889} & \underline{0.800} & \underline{0.903} & \underline{0.924} \\
\bottomrule
\end{tabular}
\caption{Benchmark results of different models for the text-based molecular design task using reference descriptions. The best results are highlighted in underline for each model.}
\label{tab:round-trip_with_gt}
\end{table*}
\paragraph{Training Details and Metrics.}
LLMs are trained following the process described in Section Method. We evaluate model performance using (1) Chemical metrics focus on assessing the quality of molecular outputs. We use the same metrics defined in Eq.~\ref{eq:metric}. (2) Textual metrics assess the quality of generated natural language descriptions, and we can also apply them to molecular outputs. We report BLEU and METEOR, two widely-used generation metrics that measure $n$-gram overlap and semantic similarity between the generated and reference texts. 
Additional implementation details and prompts are provided in Appendix~E.

\subsection{Round-trip Molecular Captioning}

Table~\ref{tab:round-trip} presents round-trip evaluation results. General-purpose LLMs like GPT-4o and Gemini-2.5-Flash perform poorly on chemical structure recovery metrics (MACCS, RDK, Morgan), indicating that strong general reasoning capabilities fail to ensure reliable molecular understanding in generative tasks. 

Crucially, integrating \model consistently enhances performance across all backbones. With ChemDFM, exact match improves by roughly 47\%, validity by 9\%, and Morgan similarity by 31\%. Similar gains are observed for ChemT5 and Qwen, demonstrating that \model is model-agnostic and broadly effective. 
These improvements extend beyond chemical metrics to textual alignment (BLEU, METEOR), reflecting enhanced bidirectional molecule-text understanding.

Among the three backbones evaluated, ChemDFM achieves the largest performance gains, followed by Qwen3, with ChemT5 showing the least improvement. This trend can be attributed to the characteristics of each model.
Qwen3, being a general-purpose language model, benefits from the introduction of chemical rewards and round-trip training but remains limited by its lack of domain-specific knowledge, resulting in comparatively lower overall performance.
ChemT5, in contrast, is domain-specific but significantly smaller (0.2B parameters). Its limited capacity leads to overfitting on the available datasets and leaves less room for improvement under our framework.
ChemDFM combines a large model size with extensive pre-training on chemical data, making it well-suited to our round-trip optimization. As a result, it not only achieves the highest baseline performance but also exhibits the largest gains when enhanced with our method.

\begin{figure*}[h!]
    \centering
    \includegraphics[width=0.95\linewidth]{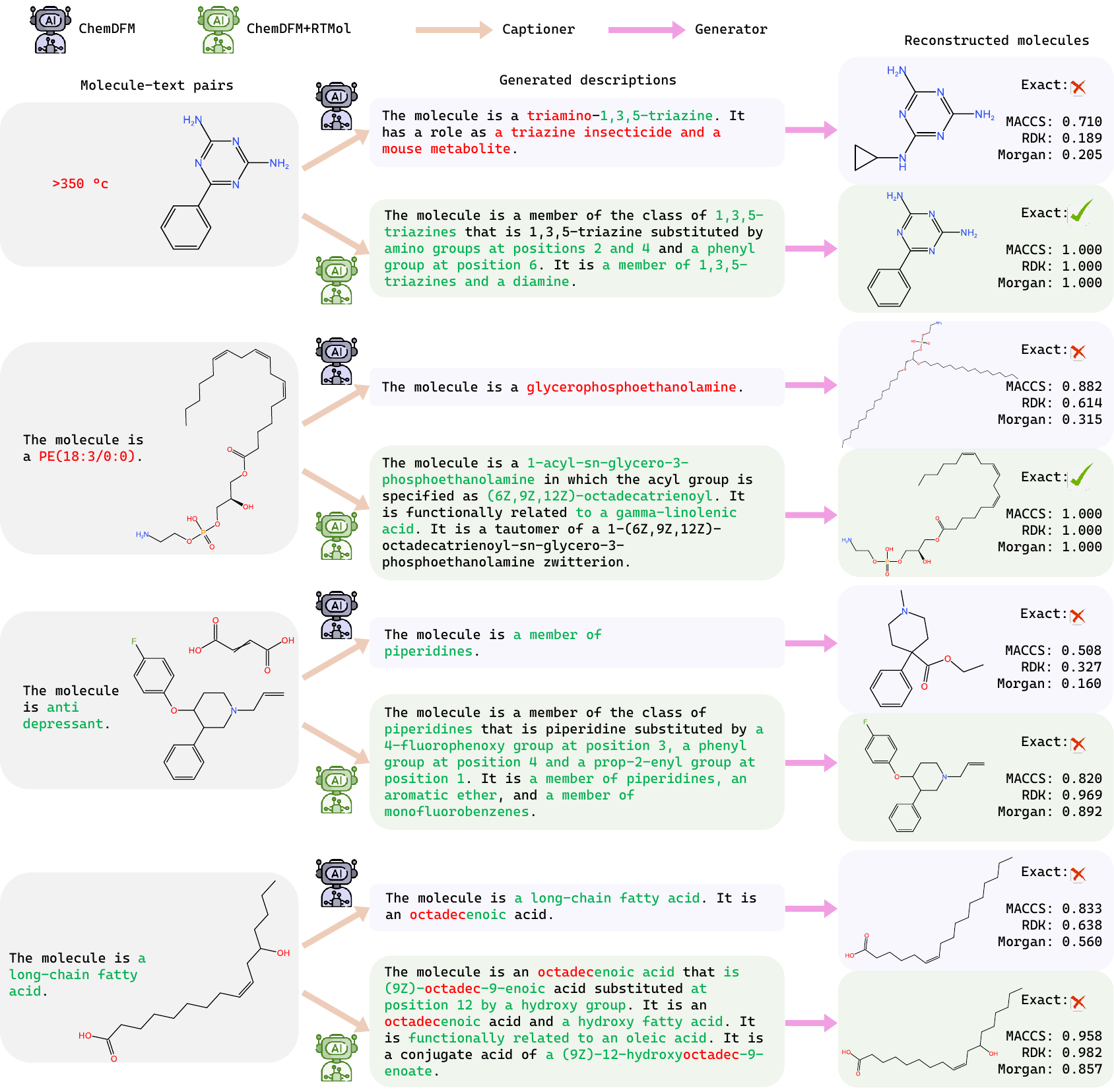}
    \caption{Cases of round-trip evaluation from the L+M-F and Mol-Instruct-F datasets. Mistakes and corrections are highlighted in red and green, respectively.}

    \label{fig:case}
\end{figure*}

These findings highlight the value of round-trip evaluation as a more faithful measure of caption quality than traditional text-only metrics. Cases in Fig.~\ref{fig:case} further illustrate how \model promotes chemically accurate and semantically informative descriptions.

\subsection{Reference Text-based Molecular Design}
We further evaluate all models using reference molecular descriptions from the ChEBI-20 dataset (Table~\ref{tab:round-trip_with_gt}). Compared to round-trip generation, performance generally improves, reflecting the benefit of high-quality textual input. For example, ChemDFM+\model achieves an 18\% increase in RDK similarity. The only exception is ChemT5, which shows marginal change. 
Despite these gains, most LLMs still struggle to fully recover molecular SMILES from text. Chemical and textual scores remain modest, with validity below 90\% across baselines. This highlights the persistent challenge of achieving precise molecular understanding and alignment, even when reference descriptions are clean and informative.

Applying our proposed \model consistently improves results across all backbones. Qwen3 benefits noticeably, while ChemDFM shows the strongest improvement, with exact match and Morgan similarity rising by 125\% and 29\%, respectively. These results show that round-trip training complements LLMs, confirming the effectiveness of our self-supervised alignment strategy. Finally, we note that such evaluations depend on the quality and coverage of reference captions. While ChEBI-20 offers reliable descriptions, other datasets remain noisy or underspecified. We analyze the impact of description quality in the next section and present cross-dataset ablation studies to demonstrate the robustness of \model with less informative descriptions.

\begin{table*}[h!]
\centering
\begin{tabular}{@{}lccccccc@{}}
\toprule
\textbf{Dataset} & \textbf{Exact(\%)}\up & \textbf{Validity(\%)}\up & \textbf{MACCS}\up & \textbf{RDK}\up & \textbf{Morgan}\up & \textbf{BLEU}\up & \textbf{METEOR}\up \\
\midrule
\multicolumn{1}{l}{L+M-F} & \multicolumn{5}{c}{\textit{Chemical metrics}} & \multicolumn{2}{c}{\textit{Textual metrics}} \\
\midrule
reference & $<$0.1 & 97.4  & 0.378  & 0.231  & 0.142  & 0.150  & 0.211 \\
generated & \underline{2.7} & \underline{99.9}  & \underline{0.892}  & \underline{0.792}  & \underline{0.624}  & \underline{0.708}  & \underline{0.750} \\
\midrule
\multicolumn{1}{l}{Mol-Instruct-F} & \multicolumn{5}{c}{\textit{}} & \multicolumn{2}{c}{\textit{}} \\
\midrule
reference & \underline{4.3} & 97.8  & 0.604  & 0.423  & 0.196  & 0.288  & 0.471 \\
generated & 2.6 & \underline{99.9}  & \underline{0.676}  & \underline{0.486}  & \underline{0.245}  & \underline{0.358}  & \underline{0.520} \\
\bottomrule
\end{tabular}
\caption{Molecular design task using reference and generated descriptions on L+M-F and Mol-Instruct-F datasets with ChemDFM+\model. The best results are highlighted in underline for each dataset.}
\label{tab:generate_data}
\end{table*}

\begin{table*}[h!]
\small
\centering
\begin{tabular}{@{}llccccccc@{}}
\toprule
\textbf{Model} & \textbf{Dataset} & \textbf{Exact(\%)}\up & \textbf{Validity(\%)}\up & \textbf{MACCS}\up & \textbf{RDK}\up & \textbf{Morgan}\up & \textbf{BLEU}\up & \textbf{METEOR}\up \\
\midrule
\multicolumn{1}{l}{ChemDFM} & & \multicolumn{5}{c}{\textit{Chemical metrics}} & \multicolumn{2}{c}{\textit{Textual metrics}} \\
\midrule
without RT & ChEBI-20 & \underline{30.0} & 94.0  & 0.791 & 0.702 & 0.571 & 0.712 & 0.801 \\
text. with RT & ChEBI-20 & 26.0 & 96.0  & 0.814  & 0.720  & 0.583  & \underline{0.735}  & \underline{0.823} \\
chem. with RT (ours) & ChEBI-20 & 28.0 & \underline{98.0}  & \underline{0.826} & \underline{0.729} & \underline{0.597} & 0.722 & 0.812 \\
\midrule
without RT & L+M-F & 1.2 & 96.6  & 0.839  & 0.741  & 0.577  & 0.669  & \underline{0.721} \\
with RT (ours) & L+M-F & \underline{1.3} & \underline{99.8}  & \underline{0.860}  & \underline{0.778}  & \underline{0.620}  & \underline{0.674}  & 0.720 \\
\midrule
without RT & Mol-Instruct-F & \underline{2.1} & 82.7  & 0.513  & 0.360  & 0.162  & \underline{0.348}  & \underline{0.517} \\
with RT (ours) & Mol-Instruct-F & 1.3 & \underline{99.5}  & \underline{0.615}  & \underline{0.435}  & \underline{0.202}  & 0.328  & 0.498 \\
\bottomrule
\end{tabular}
\caption{Ablation in round-trip evaluation using generated descriptions. The best results are highlighted in underline for each dataset.}
\label{tab:ablation}
\end{table*}

\subsection{Addressing Noisy Datasets through Unsupervised Captioning}\label{sec:data_generation}

Current molecular captioning approaches rely heavily on supervised learning with paired datasets, making performance vulnerable to annotation quality. While curated datasets like ChEBI-20 provide accurate molecule-text alignments, others like L+M-F and Mol-Instruct-F contain substantial noise, ambiguity, and generic descriptions. This quality gap severely impacts downstream performance: Table~\ref{tab:generate_data} (``reference'' rows) shows that captions from these noisy datasets yield poor reconstruction for ChemDFM+\model (e.g., L+M-F exact match $<$0.1\%, Morgan similarity $<$0.15), indicating insufficient structural information.

Our unsupervised approach overcomes this limitation by generating captions without relying on reference texts. As shown in Table~\ref{tab:generate_data} (``generated'' rows), it yields substantial improvements on L+M-F, with exact match rate increasing by over 2.5 and fingerprint similarities rising by more than 0.45 across all metrics. Similar gains are observed on both chemical and textual metrics on both datasets, demonstrating its effectiveness and robustness across diverse datasets.

Case studies (Fig.~\ref{fig:case}) demonstrate this qualitatively. Reference captions are often vague or even incorrect. With \model training, the model can produce accurate and detailed descriptions of input molecules and successfully reconstruct the original molecules (first and second cases). In other scenarios, it generates generally accurate descriptions that, while not exact, still enable reconstruction of molecules with high structural similarity to the input. 
In the third case, our method accurately reconstructs the mixture's principal active component, which is the key ingredient in antidepressant formulations. In the fourth case, we recover a long-chain fatty acid with precise substituent positions and close physicochemical properties. Both cases achieve similarity above 0.8 across all three descriptors, with the RDKit fingerprint yielding particularly high values ($>$0.95).

Unlike prior works, our method also provides a diagnostic application: the round-trip metric identifies high-confidence captions within noisy datasets (L+M-F, Mol-Instruct-F), enabling dataset curation (see Appendix~F). This demonstrates our approach's dual role in both \textit{improving generation} and \textit{evaluating dataset quality}.

\subsection{Ablation Studies}
To assess the contribution of each component in \model, we perform ablation studies under two configurations: 
(1) Reward signal: Replace the chemically grounded rewards (similarities) with purely textual metrics (BLEU and METEOR). 
(2) Round-trip objective: Remove the round-trip consistency loss and train only with unidirectional supervision (text-to-molecule and molecule-to-text). We then compare performance across multiple datasets to assess the impact of this objective. Results in Table~\ref{tab:ablation} reveal two key findings:

\textbf{Importance of chemical rewards.} Substituting chemical similarities with textual metrics consistently degrades chemical fidelity (e.g., Morgan similarity drops about 2.3\% on ChEBI-20). While BLEU and METEOR scores remain comparable, chemically aware rewards better capture structural correctness, underscoring their necessity for molecule-text alignment.

\textbf{Effect of round-trip consistency.} Removing the round-trip objective leads to lower validity and similarity across various datasets. For instance, validity decreases by approximately 4.1\%, 3.2\%, and 16.9\% on ChEBI-20, L+M-F, and Mol-Instruct-F, respectively. Similarly, Morgan similarity drops by about 4.4\%, 6.9\%, and 19.8\% on these datasets. This confirms that enforcing cycle consistency improves bidirectional alignment and yields more accurate reconstructions.

Overall, both chemically grounded rewards and round-trip consistency are indispensable components for achieving robust, generalizable alignment between molecules and text.

\section{Conclusion}

We present \model, a reinforcement learning framework that unifies molecular captioning and text-based molecular design through round-trip consistency. Unlike prior approaches that train these tasks separately and rely on text-centric evaluation, \model directly optimizes for chemical fidelity by rewarding accurate reconstruction. This formulation addresses noisy captions, misleading metrics, and fragmented bidirectional alignment. Experiments on multiple datasets demonstrate that \model consistently improves round-trip scores while generating higher-quality molecule-text pairs, demonstrating its adaptability to diverse pre-trained language models. In future work, we plan to extend \model to multi-modal chemical data (e.g., 3D structures, spectra) and downstream tasks such as drug discovery and reaction planning.

\section*{Acknowledgment}
This work was supported by the National Natural Science Foundation of China (No. 62272300).

\bibliography{aaai2026}

@article{li2024molreflect,
  title={MolReFlect: Towards In-Context Fine-grained Alignments between Molecules and Texts},
  author={Li, Jiatong and Liu, Yunqing and Liu, Wei and Le, Jingdi and Zhang, Di and Fan, Wenqi and Zhou, Dongzhan and Li, Yuqiang and Li, Qing},
  journal={arXiv preprint arXiv:2411.14721},
  year={2024}
}

@article{zhao2024chemdfm,
  title={ChemDFM: A Large Language Foundation Model for Chemistry},
  author={Zhao, Zihan and Ma, Da and Chen, Lu and Sun, Liangtai and Li, Zihao and Xia, Yi and Chen, Bo and Xu, Hongshen and Zhu, Zichen and Zhu, Su and others},
  journal={arXiv preprint arXiv:2401.14818},
  year={2024}
}

@article{zhang2024unimot,
  title={Unimot: Unified molecule-text language model with discrete token representation},
  author={Zhang, Juzheng and Bian, Yatao and Chen, Yongqiang and Yao, Quanming},
  journal={arXiv preprint arXiv:2408.00863},
  year={2024}
}

@article{guo2023can,
  title={What can large language models do in chemistry? a comprehensive benchmark on eight tasks},
  author={Guo, Taicheng and Nan, Bozhao and Liang, Zhenwen and Guo, Zhichun and Chawla, Nitesh and Wiest, Olaf and Zhang, Xiangliang and others},
  journal={Advances in Neural Information Processing Systems},
  volume={36},
  pages={59662--59688},
  year={2023}
}

@inproceedings{edwards2021text2mol,
  title={Text2mol: Cross-modal molecule retrieval with natural language queries},
  author={Edwards, Carl and Zhai, ChengXiang and Ji, Heng},
  booktitle={Proceedings of the 2021 Conference on Empirical Methods in Natural Language Processing},
  pages={595--607},
  year={2021}
}

@inproceedings{molt5,
  title={Translation between Molecules and Natural Language},
  author={Edwards, Carl and Lai, Tuan and Ros, Kevin and Honke, Garrett and Cho, Kyunghyun and Ji, Heng},
  booktitle={2022 Conference on Empirical Methods in Natural Language Processing, EMNLP 2022},
  pages={375--413},
  year={2022},
  organization={Association for Computational Linguistics (ACL)}
}

@article{batgpt,
  title={BatGPT-Chem: A Foundation Large Model For Retrosynthesis Prediction},
  author={Yang, Yifei and Shi, Runhan and Li, Zuchao and Jiang, Shu and Lu, Bao-Liang and Yang, Yang and Zhao, Hai},
  journal={arXiv preprint arXiv:2408.10285},
  year={2024}
}

@article{incontextmol,
  title={Large language models are in-context molecule learners},
  author={Li, Jiatong and Liu, Wei and Ding, Zhihao and Fan, Wenqi and Li, Yuqiang and Li, Qing},
  journal={IEEE Transactions on Knowledge and Data Engineering},
  year={2025},
  publisher={IEEE}
}

@article{momu,
  title={A molecular multimodal foundation model associating molecule graphs with natural language},
  author={Su, Bing and Du, Dazhao and Yang, Zhao and Zhou, Yujie and Li, Jiangmeng and Rao, Anyi and Sun, Hao and Lu, Zhiwu and Wen, Ji-Rong},
  journal={arXiv preprint arXiv:2209.05481},
  year={2022}
}

@article{moleculestm,
  title={Multi-modal molecule structure--text model for text-based retrieval and editing},
  author={Liu, Shengchao and Nie, Weili and Wang, Chengpeng and Lu, Jiarui and Qiao, Zhuoran and Liu, Ling and Tang, Jian and Xiao, Chaowei and Anandkumar, Animashree},
  journal={Nature Machine Intelligence},
  volume={5},
  number={12},
  pages={1447--1457},
  year={2023},
  publisher={Nature Publishing Group UK London}
}

@inproceedings{molca,
  title={MolCA: Molecular Graph-Language Modeling with Cross-Modal Projector and Uni-Modal Adapter},
  author={Liu, Zhiyuan and Li, Sihang and Luo, Yanchen and Fei, Hao and Cao, Yixin and Kawaguchi, Kenji and Wang, Xiang and Chua, Tat-Seng},
  booktitle={Proceedings of the 2023 Conference on Empirical Methods in Natural Language Processing},
  pages={15623--15638},
  year={2023}
}

@inproceedings{3dmolm,
  title={Towards 3D Molecule-Text Interpretation in Language Models},
  author={Li, Sihang and Liu, Zhiyuan and Luo, Yanchen and Wang, Xiang and He, Xiangnan and Kawaguchi, Kenji and Chua, Tat-Seng and Tian, Qi},
  booktitle={The Twelfth International Conference on Learning Representations},
year={2024}
}

@article{ether0,
  title={Training a Scientific Reasoning Model for Chemistry},
  author={Narayanan, Siddharth M and Braza, James D and Griffiths, Ryan-Rhys and Bou, Albert and Wellawatte, Geemi and Ramos, Mayk Caldas and Mitchener, Ludovico and Rodriques, Samuel G and White, Andrew D},
  journal={arXiv preprint arXiv:2506.17238},
  year={2025}
}

@inproceedings{3dmolt5,
  title={3D-MolT5: Leveraging Discrete Structural Information for Molecule-Text Modeling},
  author={Qizhi Pei and Lijun Wu and Kaiyuan Gao and Jinhua Zhu and Rui Yan},
  booktitle={International Conference on Learning Representations},
  year={2024},
}

@inproceedings{chemt5,
  title = 	 {Unifying Molecular and Textual Representations via Multi-task Language Modelling},
  author =       {Christofidellis, Dimitrios and Giannone, Giorgio and Born, Jannis and Winther, Ole and Laino, Teodoro and Manica, Matteo},
  booktitle = 	 {Proceedings of the 40th International Conference on Machine Learning},
  pages = 	 {6140--6157},
  year = 	 {2023},
  volume = 	 {202},
  series = 	 {Proceedings of Machine Learning Research},
  publisher =    {PMLR},
  url = 	 {https://proceedings.mlr.press/v202/christofidellis23a.html},
}

@article{smiles,
  title={SMILES, a chemical language and information system. 1. Introduction to methodology and encoding rules},
  author={Weininger, David},
  journal={Journal of chemical information and computer sciences},
  volume={28},
  number={1},
  pages={31--36},
  year={1988},
  publisher={ACS Publications}
}

@article{Sadeghi2024CanLL,
  title={Can large language models understand molecules?},
  author={Seyedeh Shaghayegh Sadeghi and Alan Bui and Ali Forooghi and Jianguo Lu and Alioune Ngom},
  journal={BMC Bioinformatics},
  year={2024},
  volume={25},
}

@inproceedings{mol_instruct,
title={Mol-Instructions: A Large-Scale Biomolecular Instruction Dataset for Large Language Models},
author={Yin Fang and Xiaozhuan Liang and Ningyu Zhang and Kangwei Liu and Rui Huang and Zhuo Chen and Xiaohui Fan and Huajun Chen},
booktitle={The Twelfth International Conference on Learning Representations},
year={2024},
url={https://openreview.net/forum?id=Tlsdsb6l9n}
}

@inproceedings{LM,
    title = "{L}+{M}-24: Building a Dataset for {L}anguage+{M}olecules @ {ACL} 2024",
    author = "Edwards, Carl  and
      Wang, Qingyun  and
      Zhao, Lawrence  and
      Ji, Heng",
    booktitle = "Proceedings of the 1st Workshop on Language + Molecules (L+M 2024)",
    month = aug,
    year = "2024",
    address = "Bangkok, Thailand",
    publisher = "Association for Computational Linguistics",
    url = "https://aclanthology.org/2024.langmol-1.1",
    doi = "10.18653/v1/2024.langmol-1.1",
    pages = "1--9",
}

@article{Mouchlis2021AdvancesID,
  title={Advances in De Novo Drug Design: From Conventional to Machine Learning Methods},
  author={Varnavas D. Mouchlis and Antreas Afantitis and Angela Serra and Michele Fratello and Anastasios G. Papadiamantis and Vassilis Aidinis and Iseult Lynch and Dario Greco and Georgia Melagraki},
  journal={International Journal of Molecular Sciences},
  year={2021},
  volume={22},
  url={https://api.semanticscholar.org/CorpusID:231865521}
}

@article{drawbackDataset,
  title={Instruction multi-constraint molecular generation using a teacher-student large language model},
  author={Peng Zhou and Jianmin Wang and Chunyan Li and Zixu Wang and Yiping Liu and Siqi Sun and Jianxin Lin and Longyue Wang and Xiangxiang Zeng},
  journal={BMC Biology},
  year={2024},
  volume={23},
  url={https://api.semanticscholar.org/CorpusID:268536992}
}

@article{zhang2024chemllm,
  title={Chemllm: A chemical large language model},
  author={Zhang, Di and Liu, Wei and Tan, Qian and Chen, Jingdan and Yan, Hang and Yan, Yuliang and Li, Jiatong and Huang, Weiran and Yue, Xiangyu and Ouyang, Wanli and others},
  journal={arXiv preprint arXiv:2402.06852},
  year={2024}
}

@article{chemdfmx,
  title={ChemDFM-X: towards large multimodal model for chemistry},
  author={Zhao, Zihan and Chen, Bo and Li, Jingpiao and Chen, Lu and Wen, Liyang and Wang, Pengyu and Zhu, Zichen and Zhang, Danyang and Li, Yansi and Dai, Zhongyang and others},
  journal={Science China Information Sciences},
  volume={67},
  number={12},
  pages={220109},
  year={2024},
  publisher={Springer}
}

@article{barber,
  title={The im algorithm: a variational approach to information maximization},
  author={Barber, David and Agakov, Felix},
  journal={Advances in neural information processing systems},
  volume={16},
  number={320},
  pages={201},
  year={2004}
}

@article{DeepSeekAI2024DeepSeekV3TR,
  title={DeepSeek-V3 Technical Report},
  author={DeepSeek-AI and Aixin Liu and others},
  journal={arXiv preprint arXiv:2412.19437},
  year={2024},
}

@article{openai2024gpt4,
  title   = {GPT-4 Technical Report},
  author  = {OpenAI and Josh Achiam and others},
  journal = {arXiv preprint arXiv:2303.08774},
  year    = {2024},
}

@article{Yang2025Qwen3TR,
  title={Qwen3 Technical Report},
  author={An Yang and Anfeng Li and others},
  journal={arXiv preprint arXiv:2505.09388},
  year={2025},
}

@article{Comanici2025Gemini2P,
  title={Gemini 2.5: Pushing the Frontier with Advanced Reasoning, Multimodality, Long Context, and Next Generation Agentic Capabilities},
  author={Gheorghe Comanici and Eric Bieber and others},
  journal={arXiv preprint arXiv:2507.06261},
  year={2025},
}

@article{maccs,
  author    = {Durant, Joseph L. and Leland, Burton A. and Henry, Douglas R. and Nourse, James G.},
  title     = {Reoptimization of MDL Keys for Use in Drug Discovery},
  journal   = {Journal of Chemical Information and Computer Sciences},
  year      = {2002},
  volume    = {42},
  number    = {6},
  pages     = {1273--1280},
  doi       = {10.1021/ci010132r},
}

@misc{rdkit,
  title={RDKit: Open-source cheminformatics},
  author={Greg Landrum},
  year={2024},
  note={\url{https://zenodo.org/records/12782092}}
}

@article{morgan,
  author    = {Rogers, David and Hahn, Mathew},
  title     = {Extended-Connectivity Fingerprints},
  journal   = {Journal of Chemical Information and Modeling},
  year      = {2010},
  volume    = {50},
  number    = {5},
  pages     = {742--754},
  doi       = {10.1021/ci100050t},
}

@inproceedings{meteor,
    title = "{METEOR}: An Automatic Metric for {MT} Evaluation with Improved Correlation with Human Judgments",
    author = "Banerjee, Satanjeev  and
      Lavie, Alon",
    editor = "Goldstein, Jade  and
      Lavie, Alon  and
      Lin, Chin-Yew  and
      Voss, Clare",
    booktitle = "Proceedings of the {ACL} Workshop on Intrinsic and Extrinsic Evaluation Measures for Machine Translation and/or Summarization",
    month = jun,
    year = "2005",
    address = "Ann Arbor, Michigan",
    publisher = "Association for Computational Linguistics",
    url = "https://aclanthology.org/W05-0909/",
    pages = "65--72"
}

@inproceedings{bleu,
author = {Papineni, Kishore and Roukos, Salim and Ward, Todd and Zhu, Wei-Jing},
title = {BLEU: a method for automatic evaluation of machine translation},
year = {2002},
publisher = {Association for Computational Linguistics},
address = {USA},
url = {https://doi.org/10.3115/1073083.1073135},
doi = {10.3115/1073083.1073135},
booktitle = {Proceedings of the 40th Annual Meeting on Association for Computational Linguistics},
pages = {311–318},
numpages = {8},
location = {Philadelphia, Pennsylvania},
series = {ACL '02}
}

@inproceedings{cyclegan,
  title={Unpaired image-to-image translation using cycle-consistent adversarial networks},
  author={Zhu, Jun-Yan and Park, Taesung and Isola, Phillip and Efros, Alexei A},
  booktitle={Proceedings of the IEEE international conference on computer vision},
  pages={2223--2232},
  year={2017}
}

@article{brislin1970back,
  title={Back-translation for cross-cultural research},
  author={Brislin, Richard W},
  journal={Journal of cross-cultural psychology},
  volume={1},
  number={3},
  pages={185--216},
  year={1970},
  publisher={Sage Publications Sage CA: Thousand Oaks, CA}
}

@article{he2016dual,
  title={Dual learning for machine translation},
  author={He, Di and Xia, Yingce and Qin, Tao and Wang, Liwei and Yu, Nenghai and Liu, Tie-Yan and Ma, Wei-Ying},
  journal={Advances in neural information processing systems},
  volume={29},
  year={2016}
}

@article{fang2022end,
  title={An end-to-end contrastive self-supervised learning framework for language understanding},
  author={Fang, Hongchao and Xie, Pengtao},
  journal={Transactions of the Association for Computational Linguistics},
  volume={10},
  pages={1324--1340},
  year={2022},
  publisher={MIT Press One Broadway, 12th Floor, Cambridge, Massachusetts 02142, USA~…}
}

@inproceedings{zhou2016learning,
  title={Learning dense correspondence via 3d-guided cycle consistency},
  author={Zhou, Tinghui and Krahenbuhl, Philipp and Aubry, Mathieu and Huang, Qixing and Efros, Alexei A},
  booktitle={Proceedings of the IEEE conference on computer vision and pattern recognition},
  pages={117--126},
  year={2016}
}

@inproceedings{godard2017unsupervised,
  title={Unsupervised monocular depth estimation with left-right consistency},
  author={Godard, Cl{\'e}ment and Mac Aodha, Oisin and Brostow, Gabriel J},
  booktitle={Proceedings of the IEEE conference on computer vision and pattern recognition},
  pages={270--279},
  year={2017}
}

@article{liu2021density,
  title={Density estimation using deep generative neural networks},
  author={Liu, Qiao and Xu, Jiaze and Jiang, Rui and Wong, Wing Hung},
  journal={Proceedings of the National Academy of Sciences},
  volume={118},
  number={15},
  pages={e2101344118},
  year={2021},
  publisher={National Academy of Sciences}
}

@article{liu2024evaluating,
  title={Evaluating Molecule Synthesizability via Retrosynthetic Planning and Reaction Prediction},
  author={Liu, Songtao and Zhang, Dandan and Tu, Zhengkai and Dai, Hanjun and Liu, Peng},
  journal={arXiv preprint arXiv:2411.08306},
  year={2024}
}

@article{zhang2025reasoning,
  title={Reasoning-Driven Retrosynthesis Prediction with Large Language Models via Reinforcement Learning},
  author={Zhang, Situo and Li, Hanqi and Chen, Lu and Zhao, Zihan and Lin, Xuanze and Zhu, Zichen and Chen, Bo and Chen, Xin and Yu, Kai},
  journal={arXiv preprint arXiv:2507.17448},
  year={2025}
}

@article{Bajusz2015b,
  author    = {Bajusz, D. and R\'{a}cz, A. and H\'{a}berger, K.},
  title     = {Why is Tanimoto index an appropriate choice for fingerprint-based similarity calculations?},
  journal   = {Journal of Cheminformatics},
  year      = {2015},
  volume    = {7},
  pages     = {20},
  doi       = {10.1186/s13321-015-0069-3},
  url       = {https://doi.org/10.1186/s13321-015-0069-3},
}

\newcommand{\inmain}{true}




\providecommand{\inmain}{false}  

\ifx\inmain\undefined
  \documentclass{article}
  \usepackage{xspace}
  \usepackage{graphicx}
  \usepackage{amsmath}
  \usepackage{algorithm}
  \usepackage{algorithmic}
  \usepackage{amsfonts}
  \usepackage{booktabs}
  \usepackage[margin=1.8cm]{geometry}
  \usepackage{setspace}
  \setstretch{1.5}
  \begin{document}
\fi

\ifx\inmain\undefined
  \section*{Appendix for \model: Rethinking Molecule-text Alignment in a Round-trip View}
  \addcontentsline{toc}{section}{Appendix for \model}
  \appendix
  \setcounter{section}{0}
  \renewcommand{\thesection}{\Alph{section}}
\else
  \setcounter{section}{0}
  \renewcommand{\thesection}{\Alph{section}}
    \begin{center}
    \Large \textbf{Appendix}
    \end{center}
\fi


\section{A. Related Work}\label{test}

\subsection{Molecule-text Modeling}

The concept of molecule-text modeling is first brought up by Text2Mol~\cite{edwards2021text2mol}, which introduces the ChEBI-20 dataset with pairs of molecule SMILES representations and their textual captions. MolT5~\cite{molt5} further introduces molecule-caption translation tasks to evaluate the quality of molecule-test alignment. Recent advancements in molecule-text modeling are largely driven by Large Language Models (LLMs) as they demonstrate great potential in language-related applications within scientific domains~\cite{guo2023can}. LLMs such as ChemDFM~\cite{zhao2024chemdfm}, ChemLLM~\cite{zhang2024chemllm}, and BatGPT-Chem~\cite{batgpt} are pre-trained on chemical literature and SMILES data.
These models are typically adapted for downstream tasks, such as molecule captioning and text-to-molecule generation, through supervised fine-tuning (SFT) on paired datasets. In-context learning techniques are also leveraged to help LLMs better understand molecules~\cite{incontextmol,li2024molreflect}. To overcome the limitations of purely text-based representations, many approaches incorporate richer structural information. MoMu~\cite{momu}, MoleculeSTM~\cite{moleculestm}, and MolCA~\cite{molca} combines 1D SMILES and 2D graph representations for molecule-to-text generation. 3D-MoLM~\cite{3dmolm} and 3D-MolT5~\cite{3dmolt5} leverage the 3D conformation of molecules to help LLM better understand molecules. ChemDFM-X~\cite{chemdfmx} further combines five modalities in the field of chemistry to enhance the ability of molecule understanding. Separately, ether0~\cite{ether0} leverages reinforcement learning methods to perform chemical reasoning tasks using verifiable rewards. However, the application of reinforcement learning to molecule-text alignment remains less explored. Our work aims to fill this gap by proposing a brand-new round-trip framework to systematically solve this problem.

\subsection{Round-trip Strategy}
Round-trip process, also known as cycle consistency, is a fundamental concept with broad applications across various scientific and engineering domains. Historically, this principle is employed by human translators for back-translation to verify and improve translation quality~\cite{brislin1970back}. It has since been adapted into machine learning in dual learning for machine translation~\cite{he2016dual} and to enable end-to-end self-supervised learning~\cite{fang2022end}. The applicability of cycle consistency also extends into computer vision, where it serves as a powerful constraint in tasks such as semantic alignment~\cite{zhou2016learning}, unsupervised depth estimation~\cite{godard2017unsupervised}, and unpaired image-to-image translation~\cite{cyclegan}. This principle is also leveraged in chemistry, where a ``retrosynthesis-reaction prediction'' loop is utilized to evaluate molecular synthesizability and verify retrosynthesis pathways~\cite{liu2024evaluating,zhang2025reasoning}. Beyond these empirical applications, the effectiveness of the round-trip process has been theoretically grounded in the area of density estimation~\cite{liu2021density}. 

\section{B. Proof on Round-trip Optimization}

\paragraph{Assumptions.}
We make two mild regularity assumptions.

\begin{enumerate}
    \item \textbf{Lipschitz continuity.}  
          Let $\ell_\phi(x\mid y)=\log q_\phi(x\mid y)$.
          There exists a constant $L>0$ such that
          \begin{equation}
             \bigl\lvert \ell_\phi(x\mid y)-\ell_\phi(x'\mid y)\bigr\rvert
             \;\le\;L\,d(x,x')
             \quad\forall\,x,x',y,
             \label{eq:Lip_def}
          \end{equation}
          i.e.\ $\ell_\phi(\cdot\mid y)$ is $L$--Lipschitz w.r.t.\ the
          distance $d(\cdot,\cdot)$ defined in the main text.
    \item \textbf{Random reconstruction.}  
          For a given $y$, let $x'\sim q_\phi(x'\mid y)$ be a reconstructed
          molecule sampled from the Generator.
\end{enumerate}

\paragraph{Inequality derivation.}
For any $x'$,
the Lipschitz condition in Eq.~\ref{eq:Lip_def} implies
\begin{equation}
    \ell_\phi(x\mid y)
      \;\;=\;\;\log q_\phi(x\mid y)
      \;\;\ge\;\;\ell_\phi(x'\mid y)-L\,d(x,x').
    \label{eq:single_bound}
\end{equation}
Taking the expectation over the variable
$x'\sim q_\phi(x'\mid y)$ on both sides of
Eq.~\ref{eq:single_bound} yields
\begin{equation}
   \log q_\phi(x\mid y)
   \;\;\ge\;
   \mathbb{E}_{\,q_\phi(x'\mid y)}
   \bigl[-L\,d(x,x')+\ell_\phi(x'\mid y)\bigr].
\end{equation}
Setting $\alpha:=L$ and the constant
$C:=\mathbb{E}_{\,q_\phi(x'\mid y)}\bigl[\ell_\phi(x'\mid y)\bigr]$
(which depends only on $y$) immediately gives the desired result
\begin{equation}
   \log q_\phi(x\mid y)
   \;\;\ge\;
   \mathbb{E}_{\,q_\phi(x'\mid y)}
     \bigl[-\alpha\,d(x,x')+C\bigr].
\end{equation}

\paragraph{Interpretation.}
The inequality states that, under Lipschitz continuity, the log-likelihood
assigned by the Generator to any molecule $x$ cannot decay faster than
linearly with its molecular distance from \emph{any}
sample $x'$ drawn from the same Generator.

\section{C. Reinforcement Learning Using Group Relative Policy Optimization}
In this work, we use Group Relative Policy Optimization (GRPO) as our optimization strategy.

Given a question $x$ from the dataset, we sample $G$ completions $y_1, \dots, y_G \sim \pi(\cdot|x)$. Each is assigned a reward $r_1, \dots, r_G$ and a corresponding advantage:
\begin{equation}
    A_i = \frac{r_i - \text{mean}\{r_1, \dots, r_G\}}{\text{std}\{r_1, \dots, r_G\}}.
\end{equation}

Given a single problem $x$ and a group of completions $\{y_i\}$, the per-group objective is:
\begin{equation}
\begin{aligned}
    J(\theta, x, y_1, \dots, &y_G) =\\ 
    \sum_{i=1}^{G} \frac{1}{|y_i|} \sum_{t=1}^{|y_i|} &\left\{ \text{clip}\left(\frac{\pi_\theta(y_{i,t}|x, y_{i,<t})}{\pi_{\theta_\text{old}}(y_{i,t}|x, y_{i,<t})}, A_i, \epsilon\right) \right.\\
    &- \left.\beta \hat{D}_\text{KL}[\pi_\text{ref} || \pi_r; x, y_i, z_t] \right\},
\end{aligned}
\end{equation}
where $\pi_\theta$ is the policy being optimized, $\pi_{\theta_\text{old}}$ is the policy from which we sampled rollouts, $\pi_\text{ref}$ is a reference policy, and clip is the standard PPO clip function:
\begin{equation}
    \text{clip}(r, A, \epsilon) = \min\{r \cdot A, \max\{\min\{r, 1+\epsilon\}, 1-\epsilon\} \cdot A\}.
\end{equation}

\section{D. Training Process}
The pseudo-code for the training process of RTMol is shown in Algorithm~\ref{alg:0}.

\begin{algorithm}[H]
\caption{Unified Molecule-Text Round-trip Training}\label{alg:0}
\textbf{Input:} Molecule-text pair dataset $\mathcal{D} = \{(x_i, y_i)\}$; Large language model $\mathcal{L}_{\Theta}$ with parameters $\Theta$;\\
Prompt$_{\text{captioner}}$, Prompt$_{\text{generator}}$; batch size $B$; step number $k$; rollout number $n$\\
\textbf{Output:} Updated model parameters $\Theta^\ast$\\
\vspace{-0.3cm}
\begin{algorithmic}[1]
\WHILE{not converged}
    \STATE $\mathcal{G}_{\Theta} \gets \mathcal{L}_{\Theta}$ with Prompt$_{\text{generator}}$
    \STATE $\mathcal{C}_{\Theta} \gets \mathcal{L}_{\Theta}$ with Prompt$_{\text{captioner}}$
    \FOR{$t = 1$ to $k$}
        \STATE Sample batch $\{(x_j, y_j)\}_{j=1}^B$ from $\mathcal{D}$
        \FOR{$j=1$ to $B$}
            \FOR{$s=1$ to $n$}
                \STATE $x'^{(s)}_j \sim \mathcal{G}_{\Theta}(y_j)$ \hspace{0.5em} // $x'^{(s)}_j \sim q_\phi(x'|y_j)$
                \STATE $r^{(s)}_j \gets S(x_j, x'^{(s)}_j)$
            \ENDFOR
        \ENDFOR
        \STATE $\Theta \gets$ Update $\Theta$ using GRPO with rewards $\{r_j^{(s)}\}$
    \ENDFOR
    \STATE $\Theta_0 \gets \Theta$
    \FOR{$t = 1$ to $k$}
        \STATE Sample batch $\{x_j\}_{j=1}^B$ from $\mathcal{D}$
        \FOR{$j=1$ to $B$}
            \STATE $y'_j \sim \mathcal{C}_{\Theta}(x_j)$ \hspace{0.5em} // $y'_j \sim p_\theta(y|x_j)$
            \FOR{$s=1$ to $n$}
                \STATE $x'^{(s)}_j \sim \mathcal{G}_{\Theta_0}(y'_j)$
                \STATE $r^{(s)}_j \gets S(x_j, x'^{(s)}_j)$
            \ENDFOR
        \ENDFOR
        \STATE $\Theta \gets$ Update $\Theta$ using GRPO with rewards $\{r_j^{(s)}\}$
        \STATE $\Theta_0 \gets \Theta$
    \ENDFOR
\ENDWHILE
\STATE $\Theta^* \gets \Theta$
\RETURN $\Theta^*$
\end{algorithmic}
\end{algorithm}

\begin{figure*}[h!]
    \centering
    \includegraphics[width=0.99\linewidth]{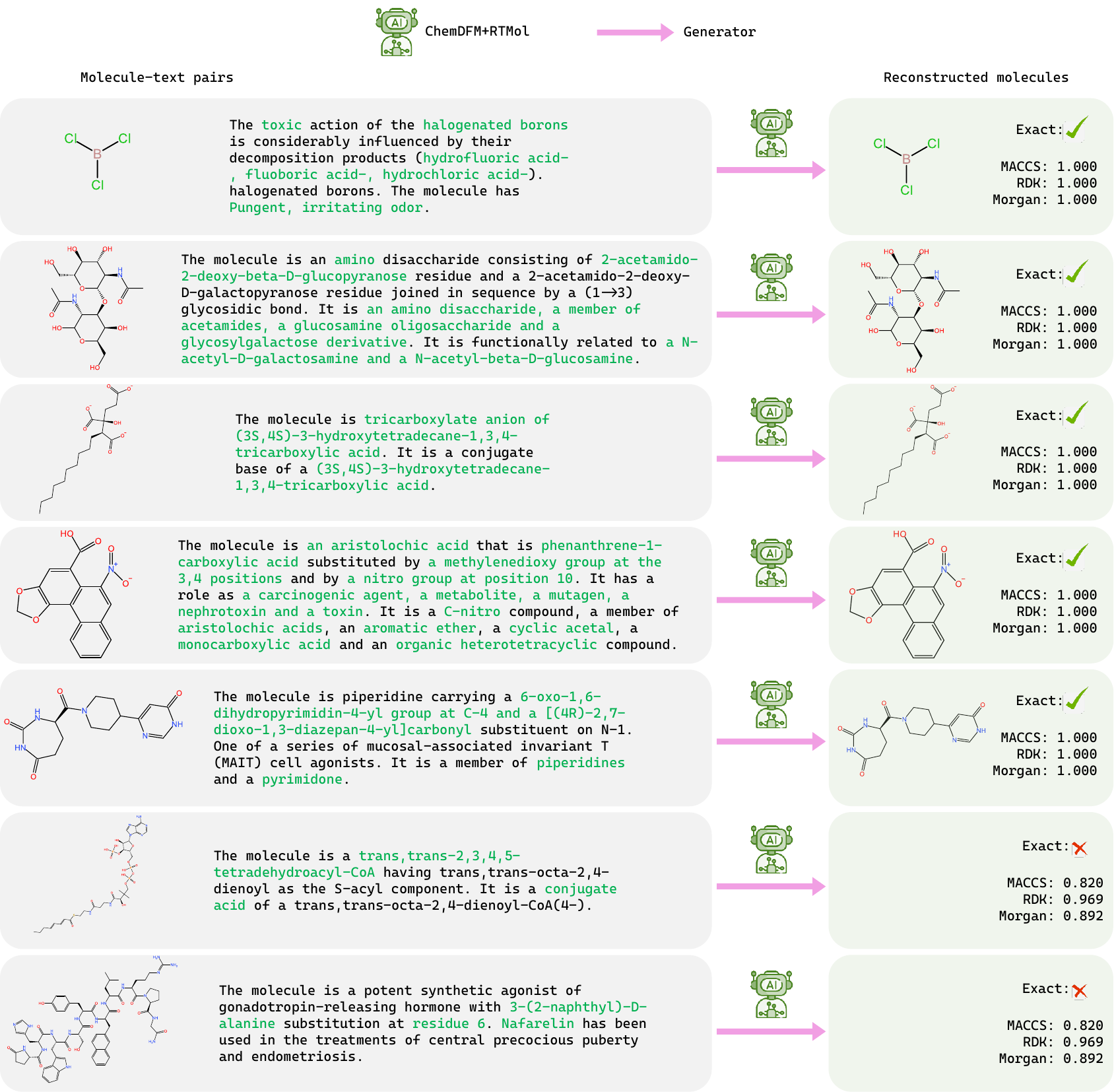}
    \caption{Examples of filtered high-quality molecule-text pairs from the L+M-F and Mol-Instruct-F datasets.}
    \label{fig:examples}
\end{figure*}
\section{E. Implementation Details}
Hyper-parameters used in \model during training are shown in Table~\ref{tab:pre-training}. We implement our framework on top of the VERL library\footnote{{https://github.com/volcengine/verl}} for GRPO training. The experiments are conducted on 8 NVIDIA A800 GPUs, requiring approximately 3, 14, and 50 hours to train ChemT5-0.2B, ChemDFM-8B, and Qwen-3-32B, respectively.
\begin{table}[htbp]
\centering
\resizebox{.98\columnwidth}{!}{
\begin{tabular}{cc|cc}
    \toprule
    Parameters & Value & Parameters & Value \\
    \midrule
    Train batch size & 128 & KL loss coefficient & 1e-3 \\
    PPO mini batch size & 64 & Learning rate & 1e-6 \\
    Rollout number & 32 & Max steps & 100 \\
    \bottomrule
\end{tabular}
}
\caption{Parameters for GRPO training.}
\label{tab:pre-training}
\end{table}

\section{F. Examples of Generated Molecule Descriptions}
Figure~\ref{fig:examples} demonstrates pairs that ChemDFM+RTMol achieves high round-trip scores.

\clearpage

\ifx\inmain\undefined
  \end{document}
\fi

\end{document}